\documentclass[conference]{IEEEtran}
\usepackage{times}

\IEEEoverridecommandlockouts

\usepackage[numbers]{natbib}
\usepackage{multicol}
\usepackage[bookmarks=true]{hyperref}

\usepackage{tablefootnote}
\usepackage{listings}
\usepackage{amsthm}
\usepackage{amsmath}
\usepackage{amssymb}
\usepackage{breqn}
\usepackage{mathtools,algorithm}
\usepackage[noend]{algpseudocode}
\usepackage{gensymb}
\usepackage{graphicx}
\usepackage{color}
\usepackage{varwidth}
\usepackage{microtype}
\usepackage{float}
\usepackage{stfloats}
\usepackage{acronym}
\usepackage{hyperref}
\hypersetup{
    colorlinks = true,
    urlcolor=magenta
}
\usepackage{multirow}
\usepackage{subcaption}
\usepackage{tabularx,booktabs}
\usepackage{lipsum}
\usepackage[utf8x]{inputenc}
\usepackage[noend]{algpseudocode}
\usepackage[symbol]{footmisc}
\usepackage[font=footnotesize]{caption} 
\definecolor{commen_green}{rgb}{0.55,0.78,0.243}

\begin{document}


\title{Vision-driven Compliant Manipulation for\\ Reliable, High-Precision Assembly Tasks}
\author{\authorblockN{Andrew S. Morgan$^{1*}$, Bowen Wen$^{2*}$, Junchi Liang$^{2}$, \\
‪Abdeslam Boularias$^{2}$, Aaron M. Dollar${^1}$, and Kostas Bekris$^{2}$}
\IEEEauthorblockA{${^1}$Deparmtent of Mechanical Engineering and Materials Science, Yale University, USA \\ Email: \tt \{andrew.morgan, aaron.dollar\}@yale.edu}
\IEEEauthorblockA{${^2}$Department of Computer Science, Rutgers University, USA\\
Email: \tt \{bw344, jl2068, ab1544, kostas.bekris\}@cs.rutgers.edu}
\thanks{This work was supported by the U.S. National Science Foundation Grants IIS-1752134 \& IIS-1900681.}
\thanks{$^*$ Authors contributed equally}
}

\maketitle
\begin{abstract}



Highly constrained manipulation tasks continue to be challenging for autonomous robots as they require high levels of precision, typically less than $1mm$, which is often incompatible with what can be achieved by traditional perception systems. This paper demonstrates that the combination of state-of-the-art object tracking with passively adaptive mechanical hardware can be leveraged to complete precision manipulation tasks with tight, industrially-relevant tolerances ($0.25mm$). The proposed control method closes the loop through vision by tracking the relative 6D pose of objects in the relevant workspace. It adjusts the control reference of both the compliant manipulator and the hand to complete object insertion tasks via within-hand manipulation. Contrary to previous efforts for insertion, our method does not require expensive force sensors, precision manipulators, or time-consuming, online learning, which is data hungry. Instead, this effort leverages mechanical compliance and utilizes an object-agnostic manipulation model of the hand learned offline, off-the-shelf motion planning, and an RGBD-based object tracker trained solely with synthetic data. These features allow the proposed system to easily generalize and transfer to new tasks and environments. This paper describes in detail the system components and showcases its efficacy with extensive experiments involving tight tolerance peg-in-hole insertion tasks of various geometries as well as open-world constrained placement tasks.

\end{abstract}

\IEEEpeerreviewmaketitle

\section{Introduction}

Developing robotic systems capable of operating in contact-rich environments with tight tolerances has remained an open research challenge. Despite progress, this is especially true for precision manipulation, where an object must be perceived, grasped, manipulated, and then appropriately placed given task requirements \cite{kemp2007,kroemer2019,correll2018analysis}. Such functionality is common for everyday insertion tasks e.g. stacking cups into one another, placing a key into a lock, or packing boxes, which are skills particularly advantageous to investigate for developing more capable robots in various application domains \cite{vandenBerg2011,bhattacharjee2019}. 


\begin{figure}[thpb]
      \centering
      \includegraphics[width=0.48\textwidth]{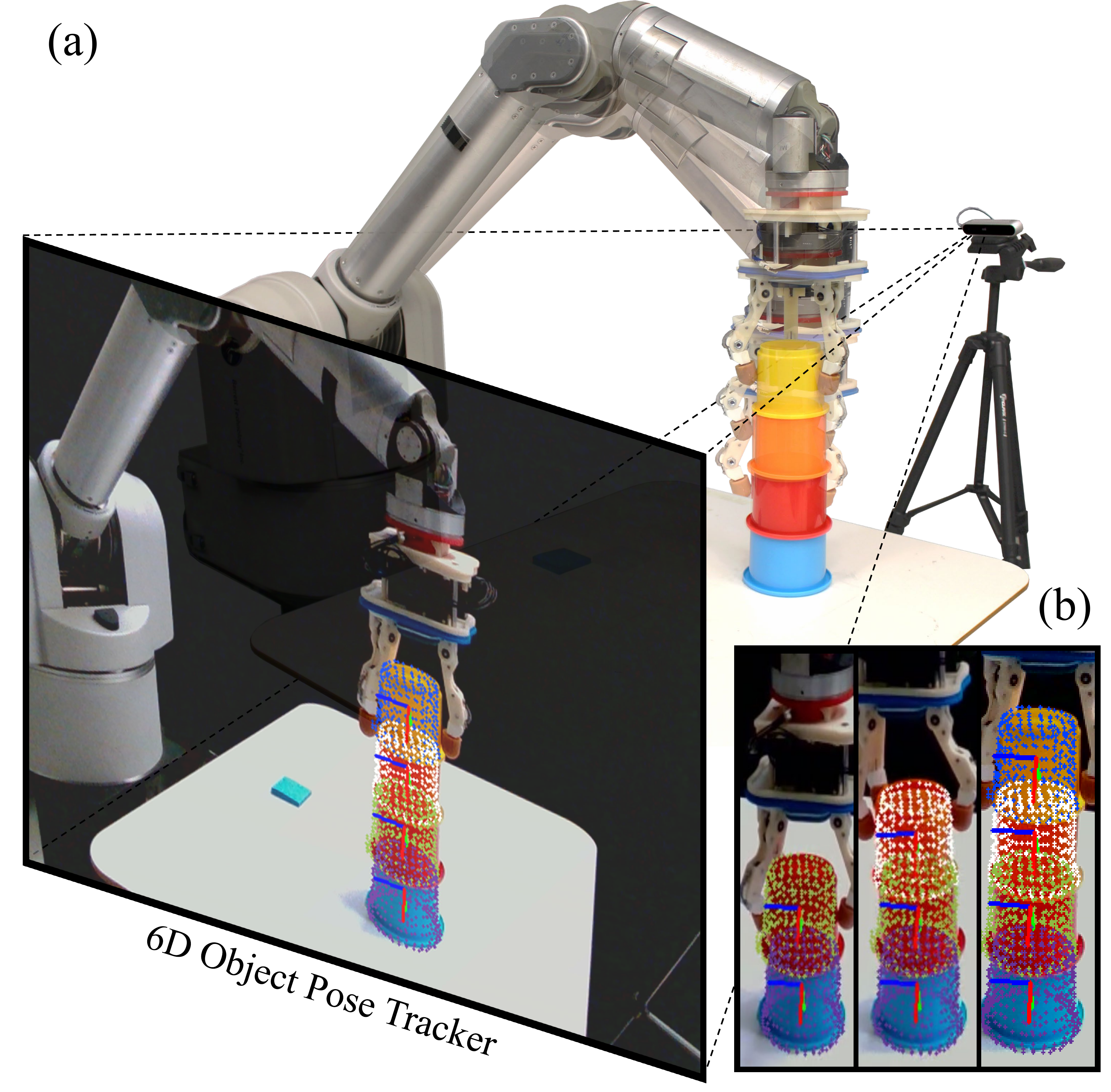}
      \vspace{-0.1cm}
      \caption{(a) An RGBD-based 6D object pose tracker monitors the task state, serving as the primary sensing modality for the robot performing a variety of insertion tasks with tight tolerances, (b) the sequence of cup stacking.}
      \label{Splash}
      \vspace{-0.55cm}
  \end{figure}


A robotic system's compliance, i.e., the adaptability of its kinematic configuration, has been key in dealing with uncertainty. Failures during manipulation are often attributed to uncertainty introduced in internal modeling, sensor readings, or robot state, which makes the system act undesirably in contact-rich scenarios, such as peg-in-hole or assembly tasks. Mechanical compliance, unlike its algorithmic counterpart \cite{hogan1984,wimbock2011}, is inherent to the robot's design and enables the system to passively reconfigure when external contact is enacted \cite{dollar2010,laliberte2002}. Although beneficial, the robot typically lacks otherwise required sensing modalities for precision control, i.e., joint encoders and force sensors can be absent. This poses a limitation, as there exists little knowledge about the robot's true state \cite{hang2020hand, vanhoof2015learning,elangovan2019}, which complicates the achievement of sub-\textit{mm} accuracy. 


Estimating the task's state dynamically, such as the object's 6D pose \cite{tjaden2018region,li2018deepim,wen2020se}, when onboard sensing is unavailable can be achieved through feedback from inexpensive, external alternatives, such as RGBD cameras \cite{chaumette2016, calli2019robust}. Although requiring additional computation, this sensing modality is advantageous as it does not require invasive, bulky sensor suites on the robot, while also providing a wider ``field of sensing'' for 
perceiving an extended workspace.

This work investigates whether a framework for compliant systems using exclusively visual feedback can perform precision manipulation tasks. It showcases a complete system and conducts numerous peg-in-hole insertions of varied geometries, in addition to completing several open world constrained placement tasks. In this way, it tests the boundaries of what is possible for vision-driven, compliance-enabled robot assembly, focusing on three main goals: 1) Perform tight tolerance insertion tasks without force sensing, a precise manipulator, or task-specific, online learning; 2) Leverage the extended workspace afforded by compliant, within-hand manipulation to increase the reliability of insertion; and, 3) Demonstrate the role of vision-based feedback for tight tolerance tasks with system compliance. To accomplish these goals, this effort brings together the following components:
\begin{itemize}
  \item \textbf{Vision-based Object Pose Tracker:} Utilizes state-of-the-art low-latency RGBD-based 6D object pose tracking \cite{wen2020se}, which is shown to be accurate enough when combined with mechanical compliance to solve the target tasks. The visual tracking monitors the task's state in the form of the object's 6D pose in a way that is robust to occlusions and varied lighting conditions, while trained solely offline on simulated data  (Sec. \ref{sec:6D tracking}).
  \item \textbf{Object-Agnostic Within-Hand Manipulation:} Uses a learned model of the inverse system dynamics of an open source and underactuated dexterous hand that is object agnostic  \cite{morgan2020}. The model is used to perform within-hand manipulation for object orientation alignment that facilitates insertion (Sec. \ref{sec: learning wihm}). 
  \item \textbf{Visual Feedback Controller: } Develops an insertion control strategy that relies exclusively on the task's state, which closes the loop through the object tracker's 6D pose estimate and generalizes to objects of differing geometries so that it is applicable to different scenarios (Sec. \ref{sec: algorithm}). The controller intentionally leverages contacts with the environment and compliance in order to increase reliability, similar to the notion of extrinsic dexterity \cite{Chavan-Dafle-2014-7860}.
\end{itemize}

This article evaluates the efficacy of the complete system by performing a variety of high-precision manipulation tasks. The results in Sec. \ref{sec:results} showcase that the proposed general object insertion algorithm, which relaxes rigid insertion constraints due to compliance, exhibits a high rate of success for tight tolerance applications. There are demonstrations indicating that the system further generalizes to a variety of everyday precision placement tasks -- stacking cups, plugging a cord into an outlet, inserting a marker into a holder, and packing boxes -- underscoring the system's practical use for complex manipulation scenarios.


\section{Related Work}
Strategies for peg-in-hole insertion, or more generally robot assembly tasks, have been studied from numerous viewpoints for several decades raising many challenges \cite{kemp2007}.  Insertion strategies for pegs of various geometries have been previously studied and attempted via a variety of possible techniques and system models: standard cylinders \cite{Park2017,tang2016,tang2016learning}, multiple-peg objects \cite{xu2019, fei2003}, soft pegs \cite{zhang2019}, industrial inserts \cite{schoettler2020}, and open world objects \cite{polverini2016,levine2015learning,choi2016vision,chang2018robotic,she2020cable}. This work seeks to generalize insertion for various geometries. 


\textbf{Model-Based Insertion:} Approaches using contact models reason about state conditions and optimal insertion trajectories while controlling the manipulator \cite{fei2003, zhang2019}. These techniques typically require expensive force-torque sensing to detect peg-hole interactions \cite{tang2016, liu2019}. Visual-based methods have been less successful, as estimation uncertainty often causes the manipulator to apply unwanted forces either damaging the robot or its environment \cite{xu2019compare}.

Compliance-enabled architectures, both hardware-based \cite{polverini2016, choi2016vision} and software-based \cite{Park2017}, are widely used to overcome uncertainties in modeling, sensing, and control of a purely rigid system. Such solutions can be applied to the manipulator, the end effector, or both. The majority of previous peg-in-hole insertion works investigate compliance in the manipulator's control -- fixing the object directly to the manipulator and evaluating trajectory search spaces or force signatures \cite{Park2017, liu2019}. Principally, these works do not address robot assembly tasks, as an end effector is vital for the completion of a pick-and-place style objective. Few works have previously investigated the role of within-hand manipulation for peg-in-hole \cite{vanwyk2018}, finding that compliance in a dexterous hand extends the workspace for task completion success. This prior work utilizes a rigid hand with tactile sensors and impedance control, which is computationally inefficient compared to the proposed framework. 

\textbf{Learning-Based Insertion: }
Regardless of whether a system is rigid or compliant, developing the control policies associated with tight tolerance tasks remain difficult \cite{xu2019compare}. To address this, learning has been widely performed on such systems -- either collecting data through robot interactions in its environment or from human-in-the-loop demonstrations \cite{tang2016learning}. Reinforcement learning and self-supervised learning \cite{lee2019making, zakka2020form2fit} have been popular, as they enable optimal policy acquisition without manual data intervention. Conceptually, this feature can be favorable for manipulation learning, as it allows the robot to sufficiently explore its environment and collect representative data of its interactions \cite{schoettler2020,levine2015learning,thomas2018learning,Beltran_Hernandez_2020}. But it is also time consuming, computationally expensive, and the extensive number of interactions needed to learn in a physical environment increases the likelihood of robot damage. It is thus advantageous to develop planning and control techniques that do not require task-specific learning or numerous physical interactions, allowing the system to more easily generalize to novel tasks and environments, as the method proposed here.

 \begin{figure*}
 \centering
 \includegraphics[width = 0.95\textwidth]{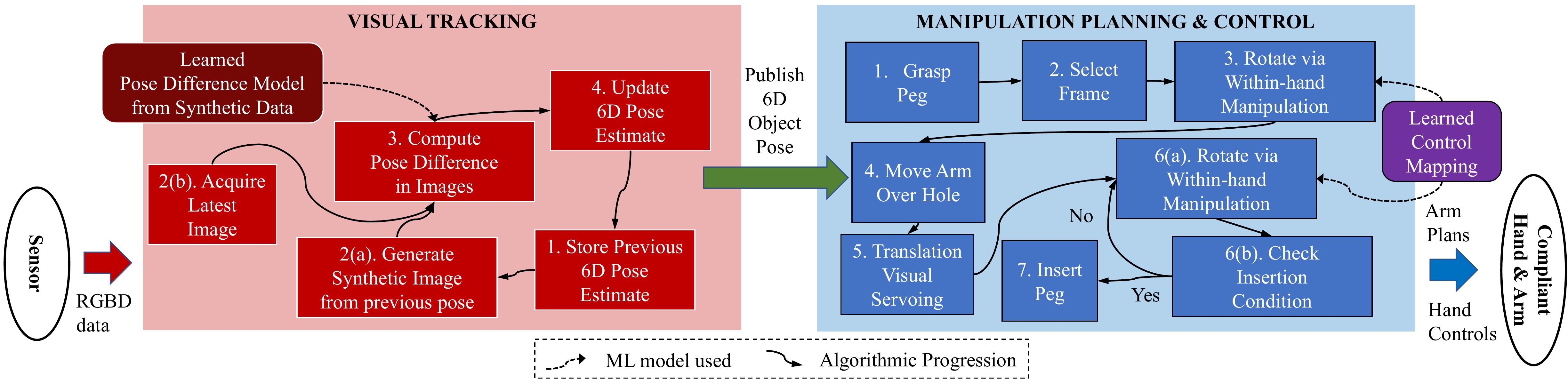}
 \caption{\textbf{System pipeline:} (red) a visual tracking framework trained solely with synthetic data to estimate 6D pose differences, provides feedback for (blue) manipulation planning and control of a low-impedance manipulator and a compliant end-effector performing within-hand manipulation for insertion tasks.}
 \vspace{-.1in}
 \label{fig:pipeline} 
 \end{figure*}

\textbf{Visual Feedback Closed-loop Manipulation: } 
Tremendous progress has been made to adapt control policies reactively with sensing feedback. With recent advances in deep learning, a number of prior works learn a visuomotor controller either by directly mapping raw image observations to control policies \cite{levine2016end,levine2018learning,viereck2017learning,andrychowicz2020learning} or by using reinforcement learning based on learned visual dynamics models \cite{ebert2018visual,bechtle2020learning,manuelli2020keypoints,byravan2018se3}. However, this method usually requires a nontrivial amount of training data which is costly to collect in the real world, struggles to transfer to new scenarios, and suffers from the curse of under-modeling or modeling uncertainties \cite{kober2013reinforcement}. Our work shares the spirit of another line of research that decouples the system into individual sensing and planning components. In particular, \cite{kappler2018real} developed a compliant manipulation system by integrating a 6D object pose tracker and a reactive motion planner. To generalize to objects of unknown shapes, \cite{mitash2020task} presents a system that maintains a dual representation of the unseen object's shape through visual tracking, achieving constrained placement. Although promising results of manipulating one or a few objects have been shown, their scalability to other objects or precision placement tasks remains unclear. In \cite{gao2021kpam}, both vision and torque feedback were used for tight insertion. Our work aims instead to robustly tackle a wide range of high precision placement tasks by leveraging the synergy between mechanical compliance and visual feedback, without force or torque feedback.

\section{Proposed Framework} \label{sec:methods}

The proposed framework integrates the following components to complete tight tolerance and open world insertion tasks: a 6D object pose tracker given RGBD data, and a compliance-enabled insertion algorithm for a passively compliant arm and dexterous hand. As depicted in the system pipeline (Fig. \ref{fig:pipeline}), during manipulation, the object tracker (red) asynchronously estimates object poses in the task space, i.e., both the peg and the hole, based on a learned pose difference model, and provides feedback for the insertion planning and control algorithm (blue) to compute action decisions based on an arm motion planner and a learned control mapping of the hand. The following subsections describe these components in more detail. 

\subsection{Visual 6D Object Pose Tracking based on Synthetic Training} \label{sec:6D tracking}


Uncertainty about the task's state can arise from multiple sources in the target application: (a) the compliant robot arm, (b) the adaptive hand, (c) the potentially occluded object, and (d) the location of the hole. Therefore, dynamic reasoning about the spatial relationship between the peg and the hole is required to achieve reliable tight insertion. This work leverages recent advances in visual tracking that employ temporal cues to dynamically update the 6D pose of tracked objects. In particular, recent work in visual tracking \cite{wen2020se} achieves robust and accurate enough estimates at a low latency to work with a wide range of objects. This allows easy integration of visual tracking with planning and control for closing the feedback loop. Additionally, it is also possible to disambiguate the 6D pose of geometrically symmetric objects from semantic textures, thanks to jointly reasoning over RGB and depth data, enabling the overall system to perform a wider range of tasks. Alternative 6D pose tracking methods, which are based solely on depth data  \cite{issac2016depth,wuthrich2013probabilistic}, often struggle with this aspect. For instance, the green charger (Fig \ref{fig:objects}) exhibits a $180\degree$ shape symmetry from the top-down view, whereas only one of the orientations can result in successfully plugging it into a power strip.


The tracker requires access to a CAD model of the manipulated object for training purposes but does not use any manually annotated data. It generalizes from synthetic data to real-world manipulation scenarios. CAD models for this purpose can be imperfect and obtained through inexpensive depth scanning processes \cite{izadi2011kinectfusion} or from online CAD libraries. The dimensional error of the CAD model can be larger than that of the tight insertion task considered -- given the adaptability afforded to the system via compliance (see plug insertion in Sec. \ref{sec:open world tasks}). At time $t$, the visual tracker operates over a pair of RGBD images $I_t$ and $I_{t-1}$ and predicts the relative pose of the tracked objects parametrized by a Lie Algebra representation $\Delta \xi \in se(3)$. The 6D object pose in the camera's frame is then recovered by $T_{t}=exp(\Delta \xi) T_{t-1}$. During both training and testing, $I_{t-1}$ is a synthetically rendered image given the pose $T_{t-1}$ and the CAD model. During training, $I_t$ is also synthetic but it is the real image during online operation of the system. Thus, a sim-to-real domain shift exists only for the current $I_t$ image. 

\begin{figure}
 \centering
\begin{subfigure}[t]{0.4\textwidth}
    \raisebox{-\height}{\includegraphics[width=\textwidth]{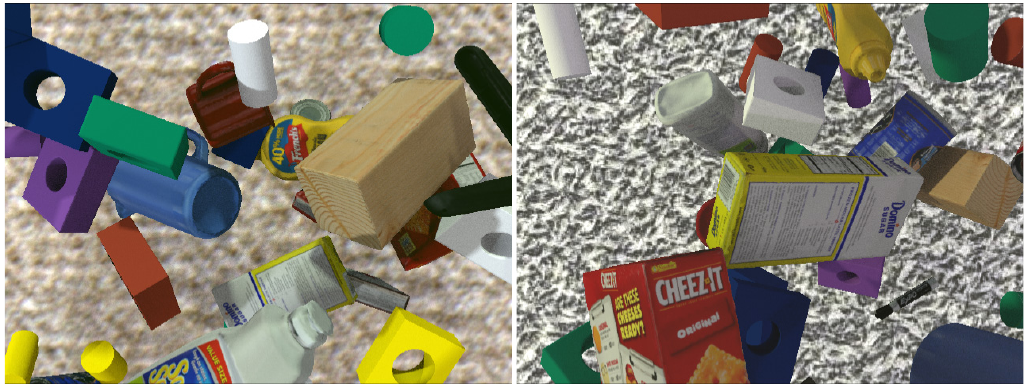}}
    \caption{}
\end{subfigure}
\begin{subfigure}[t]{0.4\textwidth}
    \raisebox{-\height}{\includegraphics[width=\textwidth]{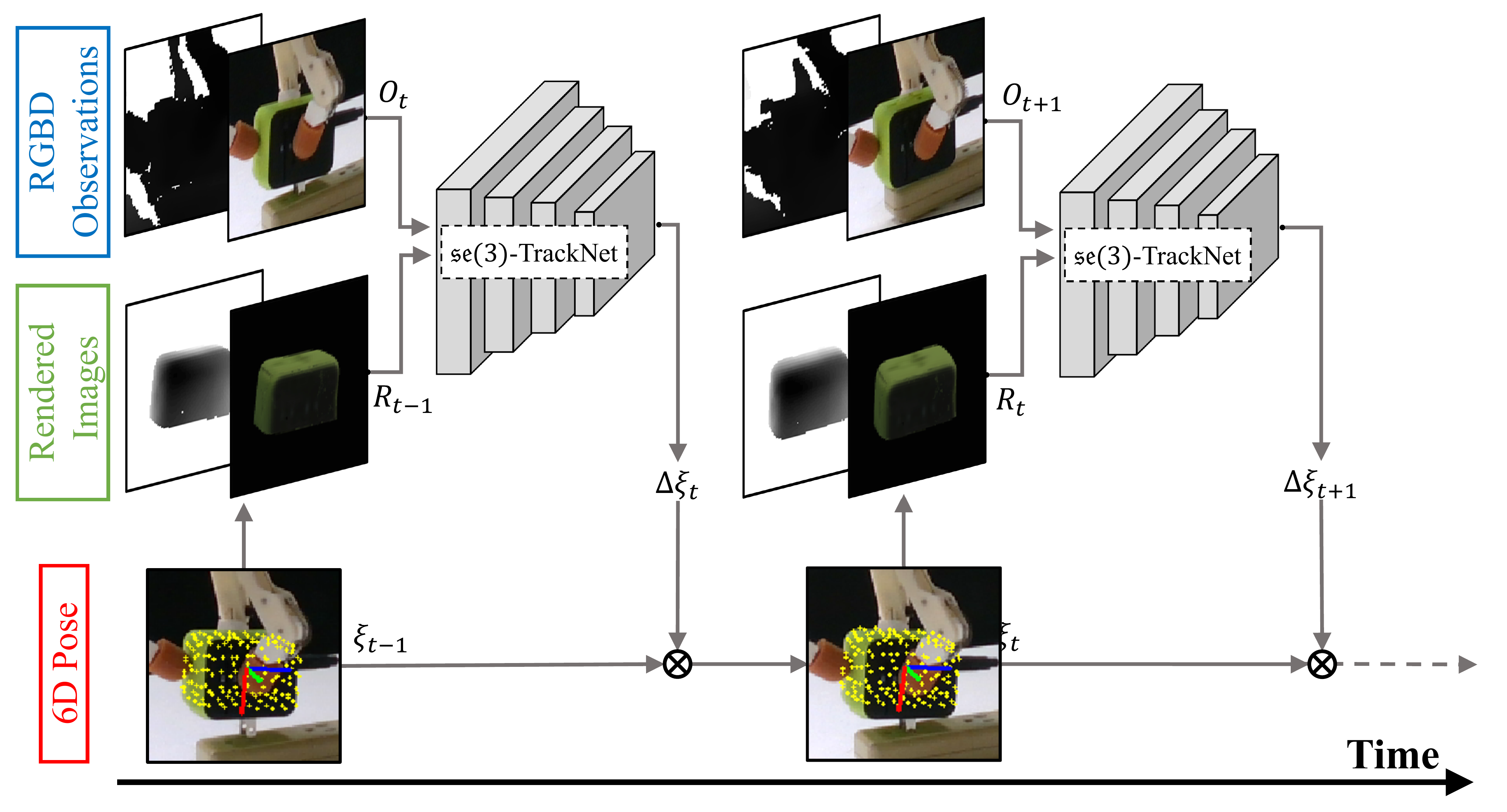}}
    \caption{}
\end{subfigure}
\vspace{-0.05in}
 \caption{(a) Physics-aware, high-fidelity synthetic training data are augmented via domain randomization. (b) 6D pose tracking on RGBD image observations streamed from the camera.}
 \vspace{-.2in}
 \label{fig:tracking}
\end{figure}

The synthetic data generation process is physics-aware and aims for generalization and high-fidelity by leveraging domain randomization techniques \cite{tobin2017domain}, as shown in Fig. \ref{fig:tracking}. To do so, external lighting positions, intensity, and color are randomized. The training process includes distractor objects from the YCB dataset \cite{calli2017} beyond the targeted objects for manipulation, which introduce occlusions and a noisy background. Object poses are randomly initialized and perturbed by physics simulation with random gravity directions until no collision or penetrations occurs. Background wall textures are randomly selected \cite{galerne2010random} for each rendering. This rendered image serves as the frame $I_{t}$. In order to generate the paired input image $I_{t-1}$ for the network, the process randomly samples a Gaussian relative motion transformation $T_{t-1}^{t}\in SE(3)$ centered on the identity relative transform to render the prior frame $I_{t-1}$. During training, data augmentations involving random HSV shift, Gaussian noise, Gaussian blur, and depth-sensing corruption are applied to the RGB and depth data in frame $I_{t}$, following bi-directional domain alignment techniques \cite{wen2020se}. Training on synthetic data takes 250 epochs and is readily applicable to real world scenarios without fine-tuning.
 

During the task execution, the proposed process tracks all manipulated objects at $30Hz$, and provides 6D poses for the planning and control module (Fig. \ref{fig:pipeline}). Before grasping, the initial 6D pose of each object on the support surface is estimated once via \textit{RANSAC}-based plane fitting and removal \cite{rusu20113d}, followed by a single-shot pose estimation approach \cite{wen2020robust} to initialize tracking. The tracking process is robust to occlusions and does not require pose re-initialization during manipulation, which is continuously executed until task completion.


\subsection{Learning Object-Agnostic Within-hand Manipulation} \label{sec: learning wihm}

\begin{figure}[thpb]
        \vspace{-.1in}
      \centering
      \includegraphics[width=0.4 \textwidth]{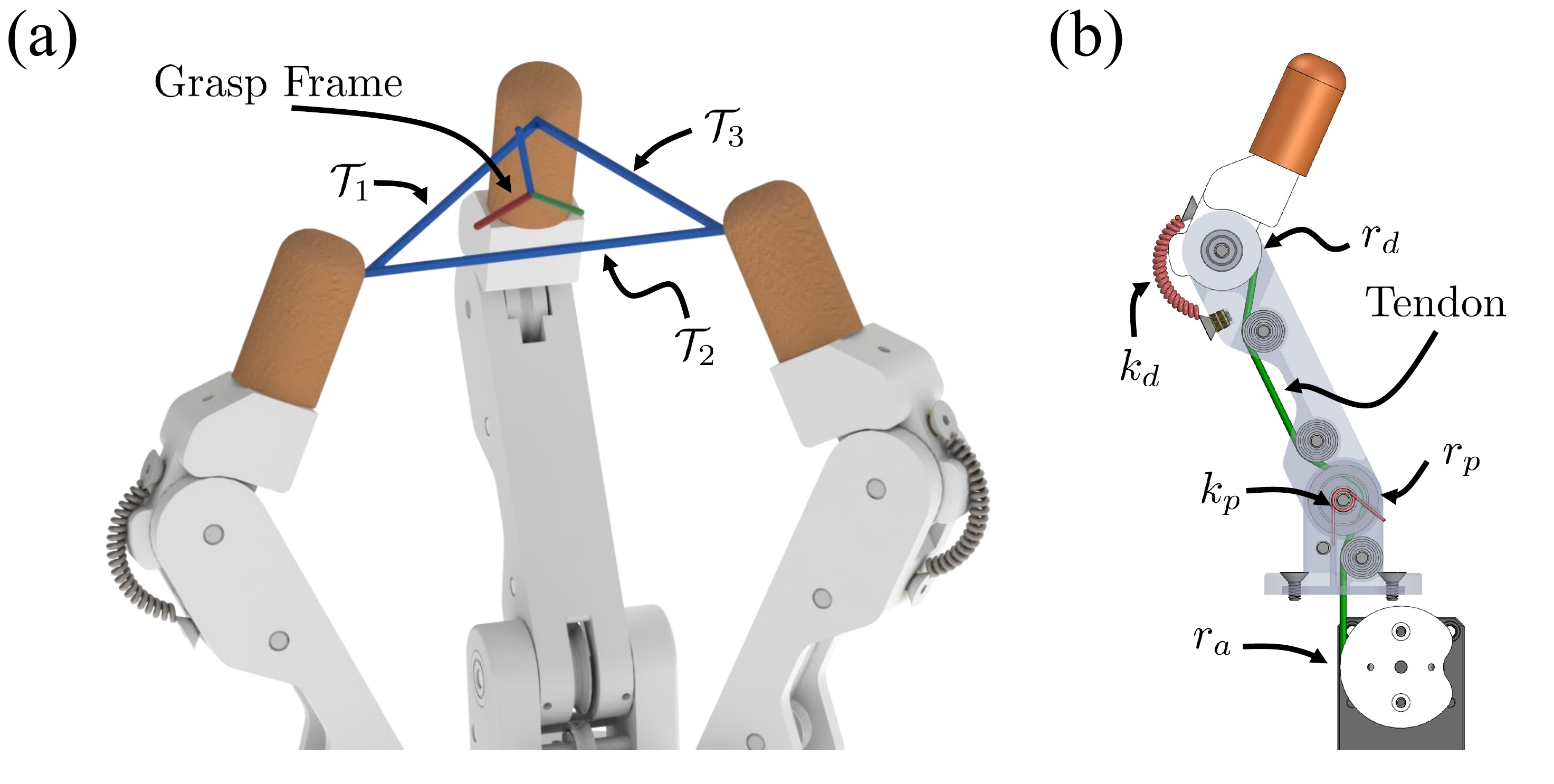}
        \vspace{-.1in}
      \caption{ (a) Object geometry can be generalized by its resultant contact triangle relationship, $\mathcal{T}$. (b) The response of a tendon-driven underactuated finger given actuation is dependent on spring constants and pulley radii.}
      \label{UAFinger}
   \end{figure}

The employed within-hand manipulation model of a 3-fingered, tendon-driven underactuated hand described is object-agnostic. It relies solely on the reference velocities of the object to suggest manipulation actions \cite{morgan2020}. This learned model is realized by first generalizing grasp geometry -- the object is represented according to the relative pose of the fingertip contacts after a grasp is acquired. From these contacts, it is possible to strictly define an object frame, $\mathcal{X}$, according to Gram-Schmidt orthogonalization. More formally, given $k$ points of contact, $\mathcal{P} = p_1, \dots, p_k$ where $p_i \in \mathbb{R}^3, \forall i \in \{1, \dots, k\}$ between the fingertips and the object with respect to the hand frame, we use $\mathcal{P}$ to calculate $\mathcal{X}$ in a closed form \cite{tahara2010}. Notably, the relative position of contacts in $\mathcal{P}$ can sufficiently represent the local geometry of the object in its manipulation plane. From this observation, it is possible to calculate the contact triangle relationship, or distance between the fingertips,  \vspace{-0.05in}
\begin{equation} \label{triangleConstraint}
     \mathcal{T} = (||p_1 -p_2||_2,||p_2 -p_3||_2,||p_3 -p_1||_2) \in \mathbb{R}^3 \vspace{-0.05in}
 \end{equation}
 where $\mathcal{T} = (\mathcal{T}_1,\mathcal{T}_2,\mathcal{T}_3)$ (Fig. \ref{UAFinger}). Note that this representation generalizes object geometry, but not necessarily object dynamics, as the global geometry, $\Gamma$, and associated inertia terms of the object are disregarded for simplicity. 
 
Given the object frame $\mathcal{X}_t \in SE(3)$ at time $t$, it is possible to model the configuration, and thus the next-state object frame $\mathcal{X}_{t+1}$ of an underactuated system given an actuation velocity $\dot a$ based on the system's energy. With the hand's joint configuration, $q \in \mathbb{R}^{\sum_{i=1}^{k} j_{i}}$, which has $j_i$ joints per finger, the system's equilibrated joint configuration $q^*$ can be calculated such that the sum of potential energies between the fingers is minimized. As in any mechanically compliant mechanism, an underactuated system's degrees of actuation are fewer than that of its configuration, i.e., $dim(a) < dim(q)$. For the adaptive hand, the fingers are actuated via a tendon transmission routed through the joints, providing a constraint:\vspace{-0.05in}
\begin{equation} \label{tendonConstraint}
     r_{ai} \dot{a}_i = r_{pi} \dot {q}_{pi} + r_{di} \dot{q}_{di},
     \vspace{-0.05in}
 \end{equation}
where $r_{ai},r_{pi},r_{di}$ represent pulley radii of the actuator and joints in finger $i$, respectively, and $\dot{a}_i, \dot{q}_{pi}, \dot{q}_{di}$ are the velocities of the actuator and joints, respectively (Fig. \ref{UAFinger}). Given this tendon constraint, and while ensuring $\mathcal{T}_t = \mathcal{T}_{t+1}$, i.e., by maintaining integrity on the contact triangle relationship, it is possible to solve for the equilibrated joint configuration: \vspace{-0.05in}
 \begin{equation} \label{energyModel}
    q^*=\arg \min_q {\sum_{i}E^i(q^i)} \; \; s.t. \; \;\eqref{triangleConstraint}, \eqref{tendonConstraint},
    \vspace{-0.05in}
 \end{equation}
 where $E^i$ represents the potential energy in the $i^{th}$ finger: \vspace{-0.05in}
 \begin{equation}
     E^i(q^i)=\frac{1}{2}(k_p q_{pi}^2+k_d q_{di}^2).
     \vspace{-0.05in}
 \end{equation}

Through Eq. \eqref{energyModel}, which has been shown to successfully transfer physically to an underactuated hand \cite{morgan2020}, this work efficiently generates system dynamics data by varying relationships in $\mathcal{T}$ and providing random actuation velocities $\dot {a}$ to the hand. In doing so, we fill a buffer of 200k object transitions, $(\mathcal{X}_t, \dot{a}) \xrightarrow{} \mathcal{X}_{t+1}$, from 50 contact triangle relationships. By taking the element-wise difference of $\mathcal{X}_t$ and $\mathcal{X}_{t+1}$, we calculate $\dot{\mathcal{X}} \in se(3)$. Given these action-reaction pairs, we train a fully-connected network to compute the model, or partially constrained Jacobian: \vspace{-0.05in}       
    \begin{equation}\label{learnedModel}
     g:(\dot{\mathcal{X}}_x, \dot{\mathcal{X}}_y) \xrightarrow{} \dot{a}
\vspace{-0.05in}
\end{equation}
that maps the desired rotational velocity of the object about the $x-$ and $y-$axes to an actuation velocity of the hand, $\dot a \in \mathbb{R}^3$. 

\vspace{-0.2in}

\section{Insertion Strategy} \label{sec:problem formulation}
\begin{figure*}
 \centering
 \includegraphics[width = 1\textwidth]{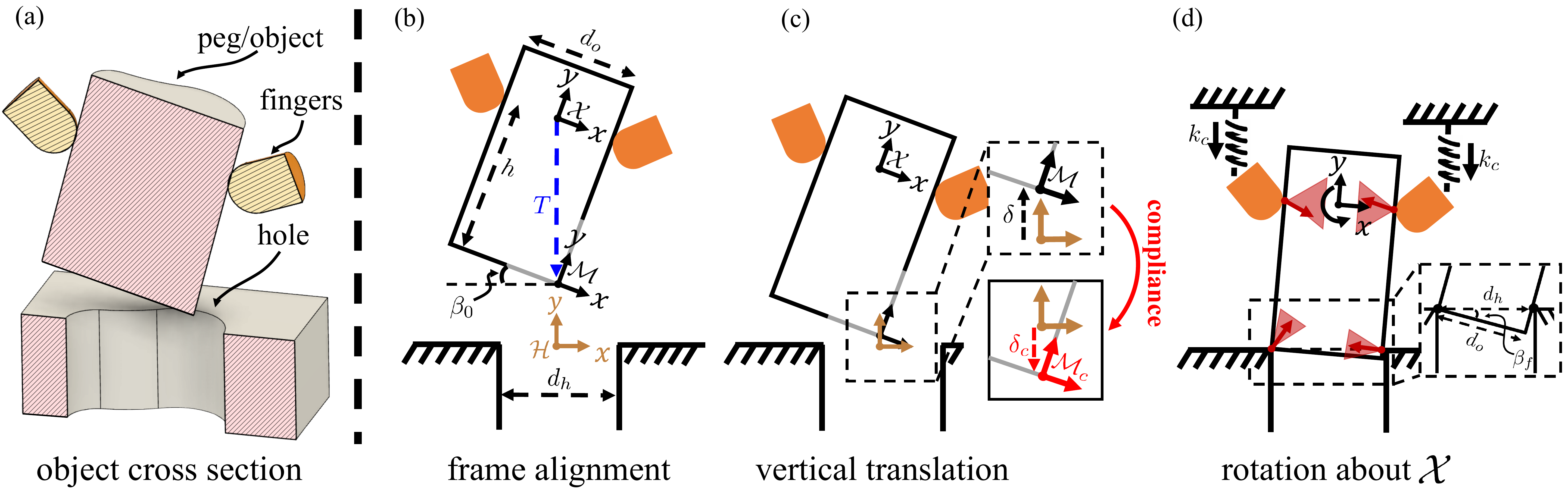}
 \caption{(a) Peg insertion is viewed as a planar challenge. (b) For insertion via purely rotational motion about $\mathcal{X}$, the peg's edge is aligned directly above the hole with a starting angle, $\beta_0$. (c) Translating $\mathcal{M}$ downward to $\delta$ guarantees that a pure rotation will align the peg with the hole, where $\delta_c$ encourages premature contact and leverages compliance to aid in alignment. (d) Upon rotation about $\mathcal{X}$, the peg must overcome contact constraints (red friction cones) of the hole contacts to align for insertion, while aided by virtual spring forces $k_c$ supplied by compliance.}
 \vspace{-.2in}
 \label{fig:Concept} 
 \end{figure*}

Assume the geometry of the peg, $\Gamma$, consists of two opposing, parallel faces with a continuous or discrete set of sidewalls connecting them, where, for the set of all antipodal point contacts on the object's sidewalls, each pair in the set has parallel contact normal force vectors. Conceptual examples of $\Gamma$ could include standard cylinders or triangular prisms. This section analyzes spatial peg insertion as a planar problem as depicted in Fig. \ref{fig:Concept},  i.e., along cross sections of the object. This approach leverages the notion that in practice we can rely on compliance and closed loop control to account for any out of plane misalignment as described in Sec. \ref{sec: algorithm}.

\subsection{The Condition for Object Insertion}\label{sec:condition for insertion}

Consider that a rigid peg is grasped such that its antipodal, i.e. set of opposing, contacts are distanced $d_o$ from one another across the width of the object. Moreover, the location of the contacts defines the height, $h$, of the grasp. Given $d_o$ and $h$, it is possible to define the object's grasp frame, $\mathcal{X} \in SE(2)$, with a pose determined by the grasp-contact vector. The task goal is to insert a peg into its corresponding rigid hole, $\mathcal{H} \in SE(2)$, which is parameterized by a width of $d_h$. Given $\Gamma$, it is possible to define a manipulation frame, $\mathcal{M} \in SE(2)$, defined by a transformation, $T$, from $\mathcal{X}$, that acts as the initial controlled frame for object insertion. Determining $\mathcal{M}$ for generalized geometries is achieved through PCA, as discussed in the accompanying \hyperref[sec:selecting a manipulation frame]{Appendix}. 

In the context of insertion tasks, this work investigates the role of object rotation via within-hand manipulation while keeping the relative translational pose between the object frame and the hole fixed. To do so, the process vertically aligns $\mathcal{H}$, $\mathcal{M}$, and $\mathcal{X}$ such that the horizontal component of $T$ is equal to $0$. This therefore sufficiently defines an initial object rotation and sets up spatial peg insertion as a planar problem with a starting angle: \vspace{-.15in} 
\begin{equation} \label{startAngle}
     \beta_0 = tan^{-1} (\frac{d_o}{2h}) \vspace{-.05in}
 \end{equation}
to satisfy the alignment constraint. It is possible to start in a different angle than $\beta_0$, but the position of $\mathcal{M}$ will need to change accordingly. Once $\beta_0$ is achieved, the system controls a vertical displacement, $\delta$, between $\mathcal{M}$ and $\mathcal{H}$, which can be calculated in closed form such that a pure rotation about $\mathcal{X}$ places the object in a state aligned with the hole. This final angle of alignment, $\beta_f$, can be calculated as: \vspace{-.05in}
\begin{equation} \label{startAngle}
 |\beta_f| \le cos^{-1} (\frac{d_o}{d_h}), \vspace{-.05in}
\end{equation}
which is dependent on the peg's two-contact case (Fig. \ref{fig:Concept}.d). Thus, the initial insertion height $\delta$ before rotation is given as \vspace{-.05in}
\begin{equation} \label{insertionDepth}
\begin{aligned}
 \delta =  h \left( cos(\beta_0) + cos(\beta_f) \right) +\frac{d_o}{2}\left(sin(\beta_0)+sin(\beta_f)\right).
 \vspace{-.05in}
\end{aligned}
\end{equation}

The sequential progression of steps for rotation-based object insertion are outlined in Fig. \ref{fig:Concept}. In summary: 1) Rotate $\mathcal{M}$ such that it aligns with $\mathcal{H}$; 2) Translate $\mathcal{M}$ downward to $\delta$; and, 3) Rotate about $\mathcal{X}$ until the two-contact case is overcome, i.e., rotation of $\mathcal{X} < |\beta_f|$, and the peg is aligned with the hole. Once satisfied, the remaining actions for insertion seek object translation downward while avoiding jamming along the sidewalls. To account for this, this work leverages compliance and controlled object spiral motions \cite{liu2019} provided by within-hand manipulation.    

\textbf{On the Role of Compliance:} While the above formulation holds for rigid insertion in the relaxed planar representation, the bounds of $\delta$ are extremely small for tight tolerance cases ($<1.5mm$ if perfectly aligned). Moreover, when considering any out-of-plane misalignment in the true additional dimension that is not modeled, the peg can easily become off-centered with respect to the hole. The features inherent to compliance benefit this task, as it not only enables adaptability to the out-of-plane misalignment, but also allows to relax these strict bounds of $\delta$ by developing a compliance-enabled insertion distance, $\delta_c$ (Fig. \ref{fig:Concept}.c). This deviation in depth is nontrivial to determine in its closed form -- as it depends on models of the manipulator-hand system's compliance, the forces that can be applied to the object via the hand, and an approximation of the nonlinear contact dynamics associated with peg-hole interactions (Fig. \ref{fig:Concept}.d). Although difficult to model, a properly tuned translational deviation in $\delta_c$ towards the hole from $\delta$ is advantageous, as it encourages contact between the hole and the object prematurely so that the contact constraints encourage and assist in alignment before insertion.

\subsection{Insertion Algorithm} \label{sec: algorithm}

The proposed system is controlled by closing the loop through visual feedback. In real time, the visual tracker monitors the state of the task space; specifically, the relative pose between the peg and the hole. This relationship is used to adjust the control reference setpoints of both the manipulator's joints and the tendons in the hand, regardless of their believed configuration states. By continually servoing the object's $SE(3)$ pose relative to the hole, it is possible to complete tight insertion tasks while adapting to external disturbances and noisy pose estimates (Sec. \ref{sec:results}). 

\textbf{Vision-Driven Object Insertion:} The insertion sequence shown in Alg. \ref{alg:mainAlgorithm} begins with the object tracker asynchronously monitoring the pose of both, the object and the hole. Once the object is grasped off of the support surface, the precomputed transformation, $T$, via PCA \eqref{PCA}, from the manipulation frame, $\mathcal{M}$, can be rigidly attached to $\mathcal{X}$. 

\vspace{-.1in}
\begin{algorithm}[htb]
	\caption{Vision-Driven Object Insertion}
	\label{alg:mainAlgorithm}
    \begin{algorithmic}[1]
    	\renewcommand{\algorithmicrequire}{\textbf{Input:}}
    	\renewcommand{\algorithmicensure}{\textbf{Output:}}
    	\Require $\Gamma, (\beta_0, \beta_f, \delta_c, \gamma, \sigma)$
    	\Comment {object geometry, $hyparams$(initial object angle, final object angle, insertion depth, alignment tolerance, step length)}
    	\State $T, \pi_1 \gets PCA(\Gamma)$
    	\Comment{POM transform, principal axis \eqref{PCA}}
	    \State $\textbf{tracker}.start\_async\_perceive(\Gamma)$
	    \Comment{start tracking thread}
	    \State $\textbf{tracker}.attach\_transform(T)$
	    \Comment {attach $T$ to $\mathcal{X}$ for $\mathcal{M}$}
	    \State $\mathcal{X}, \mathcal{H} \gets \textbf{tracker}.get\_poses()$
	    \Comment{object \& hole 6D poses}
   
    	\State $\textbf{system}.grasp(\mathcal{X}, \pi_1)$
    	\Comment{grasp and lift object}
    	
    	\State $\textbf{hand}.rotation\_servo(\textbf{tracker},\pi_1, \beta_0)$
    	\Comment{Alg. \ref{alg:manipulate}}
    	\State $\mathcal{M}, \mathcal{H} \gets \textbf{tracker}.get\_poses()$
    	\State $\textbf{arm}.move\_above(\mathcal{M},\mathcal{H})$
        \Comment{place edge above hole}
    	\State $\textbf{arm}.translation\_servo(\textbf{tracker},\delta_c, \gamma, \sigma)$ 
    	\Comment { Alg. \ref{alg:servo}}
    	\State $\textbf{hand}.rotation\_servo(\textbf{tracker},\pi_1, \beta_f)$
    	\Comment{Alg. \ref{alg:manipulate}}
    	\State $\textbf{system}.spiral\_insertion()$
    	\Comment{coordinated spiral insertion}
    	
\end{algorithmic}
\end{algorithm}

Upon doing so, within-hand manipulation is performed such that the initial insertion angle, $\beta_0$ from Sec. \ref{sec:condition for insertion} is achieved along the principal axis, $\pi_1$ (Alg. \ref{alg:manipulate}). The velocity references, $\dot{\mathcal{X}}_x$ and $\dot{\mathcal{X}}_y$, during this within-hand manipulation process are chosen so as to minimize any rotation not along this principal axis. Upon reaching $\beta_0$, the robot arm plans a trajectory such that the resultant position of the manipulation frame is directly above the hole. Due to the robot's imprecision, however, the desired pose of $\mathcal{M}$ above $\mathcal{H}$ is often not accurately achieved. 

At this point, the $translation\_servo$ method begins by sequentially adjusting the control reference of the manipulator based on feedback from the object tracker (Alg. \ref{alg:servo}). Specifically, given the hyper-parameter $\gamma$, if the hole $xy$-plane translational difference between $\mathcal{M}$ and $\mathcal{H}$ is within $\gamma$, the manipulator will move a step, $\sigma$, which can be fixed or adaptive, downward towards the hole by adjusting the joint target values of the robot. Otherwise, the robot will move in the Cartesian step, $\sigma$, towards aligning $\mathcal{M}$ with $\mathcal{H}$. This process is continually repeated until the vertical threshold, $\delta_c$, is reached.    

\vspace{-.1in}
\begin{algorithm}[htb]
	\caption{Within-Hand Rotation Visual Feedback Control}
	\label{alg:manipulate}
    \begin{algorithmic}[1]
    	\renewcommand{\algorithmicrequire}{\textbf{Input:}}
    	\renewcommand{\algorithmicensure}{\textbf{Output:}}
    	\Require \textbf{tracker}, $\pi_1$, $\theta \in \{\beta_0, \beta_f\}$
    	
    	\Comment {principal axis of rotation, object target angle}  
    	    
    	\State $\mathcal{X}, \mathcal{H} \gets \textbf{tracker}.get\_poses()$ \Comment{object and hole 6D pose}
	    \State $R_{\Delta} \gets \mathcal{X}.R - \mathcal{H}.R$
    	    \Comment{relative rotation $\in SO(3)$}    	  	    
    	\If{$\theta$ \textbf{is} $\beta_0$}
    	    \Comment{increase angle for edge insertion}
    	    \While {${\mathcal{H}.R}(\pi_1) - {\mathcal{X}.R}(\pi_1) \le \beta_0$}
    	        \State{$\dot{\mathcal{X}} \gets [-R_{\Delta}(\pi_{1_x}),-R_{\Delta}(\pi_{1_y})]$}
    	        \Comment{rotate along $\pi_1$}
        	    \State {$\dot{a} \gets \textbf{hand.model}.predict(\dot{\mathcal{X}})$}
        	    \Comment{Eq. \eqref{learnedModel}}
        	    \State $ \textbf{hand}.actuate(\dot{a})$ \Comment{send action to motors}
        	    \State $\mathcal{X}, \mathcal{H} \gets \textbf{tracker}.get\_poses()$
        	    \State $R_{\Delta} \gets \mathcal{X}.R - \mathcal{H}.R$         	    
            \EndWhile
        \EndIf
    	\If{$\theta$ \textbf{is} $\beta_f$}
    	    \Comment{decrease angle, align with hole}
    	    \While {$R_{\Delta x} \ge \beta_f$ $\textbf{or}$ $R_{\Delta y}$ $\ge \beta_f$}
    	        \State{$\dot{\mathcal{X}} \gets [R_{\Delta x}, R_{\Delta y}]$}
        	    \State {$\dot{a} \gets \textbf{hand.model}.predict(\dot{\mathcal{X}})$}
        	    \State $ \textbf{hand}.actuate(\dot{a})$
        	    \State $\mathcal{X}, \mathcal{H} \gets \textbf{tracker}.get\_poses()$
        	    \State $R_{\Delta} \gets {\mathcal{H}.R} - {\mathcal{X}.R}$
            \EndWhile
        \EndIf

\end{algorithmic}
\end{algorithm}
\vspace{-.1in}

\vspace{-.1in}
\begin{algorithm}[htb]
	\caption{Arm Translation Visual Feedback Control}
	\label{alg:servo}
    \begin{algorithmic}[1]
    	\renewcommand{\algorithmicrequire}{\textbf{Input:}}
    	\renewcommand{\algorithmicensure}{\textbf{Output:}}
    	\Require \textbf{tracker}, $\delta_c$, $\gamma$, $\sigma$

    	\Comment{insertion depth, alignment tolerance, step length}
    	
	\State {$\mathcal{M}, \mathcal{H} \gets \textbf{tracker}.get\_poses()$}
	\While{$\mathcal{M}.t_z - \mathcal{H}.t_z >\delta_c$} \Comment{check translation along z}
	    \State {$\mathcal{E}_{\Delta x},\mathcal{E}_{\Delta y},\mathcal{E}_{\Delta z} \gets 0,0,0$}
	    \If{$-\gamma \le \mathcal{M}.t_x-\mathcal{H}.t_x \le \gamma \textbf{ and } 
	    \newline \indent \indent \ -\gamma \le \mathcal{M}.t_y-\mathcal{H}.t_y \le \gamma $}
            \State {$\mathcal{E}_{\Delta z} \gets -\sigma$ }
            \Comment{ servo end effector pose downward}
	    \Else
	    
            \State $\epsilon_x \gets \mathcal{X}.t_x-\mathcal{H}.t_x$
            \State $\epsilon_y \gets \mathcal{X}.t_y-\mathcal{H}.t_y$
            \State {$\mathcal{E}_{\Delta x},\mathcal{E}_{\Delta y}  \gets -$sgn$(\epsilon_x)\sigma,$ $- $sgn$(\epsilon_y)\sigma $ }
            
            \Comment{servo translation to align with hole}
	    \EndIf
	    
        \State $\mathcal{A} \gets \textbf{arm.}motion\_planner(\mathcal{E}_{\Delta})$
        \Comment{plan motion delta}
        \State $\textbf{arm.}execute(\mathcal{A})$
         \Comment{move end effector}
         \State {$\mathcal{M}, \mathcal{H} \gets \textbf{tracker}.get\_poses()$}
	\EndWhile
\end{algorithmic}
\end{algorithm}
\vspace{-.1in}

The hand then attempts to reorient $\mathcal{X}$ with $\mathcal{H}$ while maintaining a small downward force between the object and the hole's edges, provided by system compliance and $\delta_c$.  Orientation alignment of the peg is solely achieved through within-hand manipulation (Alg. \ref{alg:manipulate}); where the object no longer follows rotations along $\pi_1$, but now by the true rotational difference $R_{\Delta}$ between the hole and the peg. This process continues until the rotational angle is less than $\beta_f$.  

The algorithm concludes by performing a spiral insertion technique along the roll and pitch axes of the object. If the object were unconstrained, the hand's actuation would provide a spiral pattern of $\mathcal{M}$, as in related work \cite{liu2019, Park2017}, while the arm attempts to slowly translate downward. This coordinated hand/arm motion helps limit jamming and encourages proper insertion until the object is fully seated into the hole. 


\section{Experiments} \label{sec:results}

We test the proposed insertion framework using a robotic system comprised of a low-impedance manipulator for object translation, an underactuated hand performing within-hand manipulation for object rotation, and a low latency 6D object pose tracker based on RGBD data (Figs. \ref{Splash}, \ref{fig:open_world_insertion}). The manipulator, a 7-DOF Barrett WAM arm, utilizes the $RRTConnect$ algorithm in OMPL \cite{sucan2012open} and is imprecise due to an inaccurate internal model of its true system dynamics, with translational errors as large as 2.6$cm$. The end effector, an adapted Yale OpenHand Model O \cite{ma2017}, is a mechanically adaptive hand comprised of six joints and three controlled actuators, and is not equipped with joint encoders or tactile sensors. Changes to the open source design include joint bearings to reduce friction, and rounded fingertips to assist in within-hand manipulation. Finally, the 6D object pose tracker, which was calibrated to the robot's world frame, takes the RGBD images streamed from an external, statically mounted Intel Realsense D415 camera, and in real-time, estimates the 6D pose of the manipulated object to provide feedback for the control loop.

\begin{figure}[ht]   
\vspace{-.1in}
	\centering
      \includegraphics[width=0.49\textwidth]{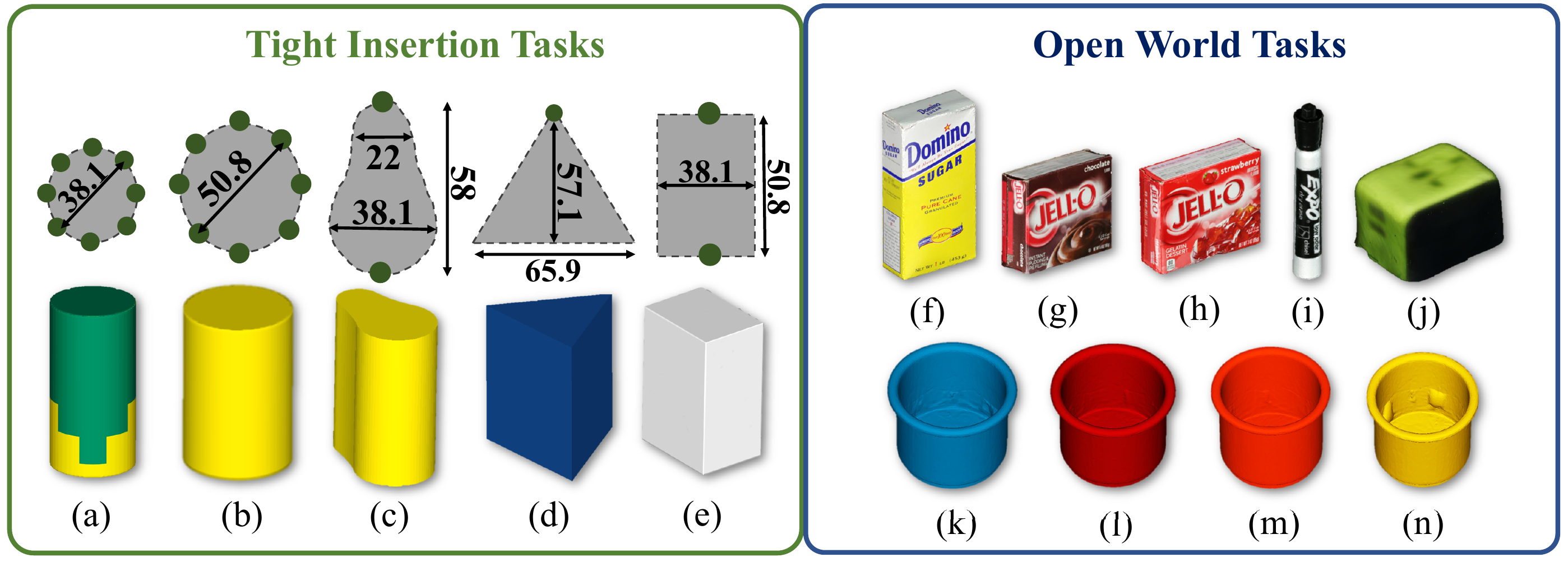}
	  \caption{Experimental objects considered in the \textbf{Tight Tolerance Tasks} and \textbf{Open World Tasks}. All lengths are in \textit{mm} and points on object faces indicate PCA-determined edge manipulation frames. (a) small circle, (b) large circle, (c) pear, (d) triangle, (e) rectangle, (f) YCB \textit{004\_sugar\_box}, (g) YCB \textit{008\_pudding\_box}, (h) YCB \textit{009\_gelatin\_box}, (i) YCB \textit{040\_large\_marker}, (j) green charger. (k)-(n) YCB \textit{065-cups}.} \vspace{-.1in}
	  \label{fig:objects}
\end{figure}

The system is tested via numerous insertion tasks. First, five 3D printed peg-in-hole objects were designed with $<0.25mm$ hole tolerances and were painted different colors and/or patterns, for evaluation on both textured and textureless objects. These objects are described based on their face geometries: namely, the small circle, large circle, pear, triangle, and rectangle (Fig. \ref{fig:objects}). Insertion was tested for each of these objects individually, where we then sequentially isolate individual system components -- compliance, control, and sensing -- to evaluate their effects on task success. Finally, we assessed the efficacy of the approach with six open world insertion tasks, involving nine different objects with diverse and challenging properties (textureless, reflective, flat and thin shapes, etc.), to underscore the utility of the framework in complex manipulation scenarios (Fig. \ref{fig:open_world_insertion}).  

\vspace{+.1in}
\begin{table}[h]
    \caption{\textsc{\small{Precision Insertion for Proposed System}}}  
    \label{tab:main insertion}
    \renewcommand{\arraystretch}{1.2}
    \centering
    \addtolength{\tabcolsep}{-1pt} 
     \begin{tabular}{c c c c c}   
     \textbf{Obj. }&\textbf{Planning (s)}&\textbf{Total (s)}&\textbf{Hand Actions}&\textbf{Success } \\[0.4ex] 
     \hline\hline
     Small Cir. & 4.1 $\pm$ 0.92 & 94.8 $\pm$ 10.78 & 48.9 $\pm$ 8.76 & \textbf{10/12}\\ 
     Large Cir. & 3.5 $\pm$ 0.52 & 88.8 $\pm$ 11.54 & 44.9 $\pm$ 9.10 & \textbf{11/12}\\ 
     Pear & 4.0 $\pm$ 1.05 & 77.9 $\pm$ 10.34 & 35.2 $\pm$ 8.62 & \textbf{9/12}\\ 
     Triangle & 3.7 $\pm$ 2.08 & 90.9 $\pm$ 14.00 & 27.4 $\pm$ 4.60 & \textbf{8/12} \\
     Rectangle & 6.2 $\pm$ 3.53 & 106.4 $\pm$ 23.43 & 40.9 $\pm$ 5.90 & \textbf{9/12} \\
     \hline
    \end{tabular}
\end{table}
\vspace{-.15in}

\subsection{Tight Tolerance Object Insertion} \label{sec:tighttolerance}
This test involved 12 insertions for each of the five objects using Alg. \ref{alg:mainAlgorithm}. Upon task reset, objects were placed back onto the support surface in no predefined pose; it was up to the system to initialize and track this pose and reacquire a grasp. The results, presented in Table \ref{tab:main insertion}, depict the planning time, total execution time, number of hand actions used for within-hand manipulation, and success rate for each object scenario. The large circle had the highest rate of success compared to any of the other objects, while the triangle had the lowest. The two objects that did not have curved edges, i.e., the triangle and the rectangle, were the most difficult out of the five to insert. The interpretation is that the constraints of the task, i.e., sharp edges of the object's face, did not enable system compliance to easily align the yaw rotation of the object with the hole. This posed a slight challenge for the relaxed planar insertion strategy from Sec. \ref{sec:condition for insertion}, and required precise yaw rotation via the manipulator, which was not required for the other objects. 

\vspace{-.1in}
\begin{table}[h]
    \caption{\textsc{\small{System Ablation Analysis}}}  
    \label{tab:ablations}
    \renewcommand{\arraystretch}{1.2}
    \centering
    \addtolength{\tabcolsep}{-3pt} 
     \begin{tabular}{c |c c c c c}   
     & \textbf{ Obj. }&\textbf{Planning (s)}&\textbf{Total (s)}&\textbf{Hand Actions}&\textbf{Success } \\[0.4ex] 
     \hline \hline
     \\ [-1.8ex]
    \multirow{12}{*}{\rotatebox[origin=c]{90}{\textbf{Reducing Compliance}}} 
     & \multicolumn{5}{c}{\textbf{Rigid Hand / Compliant Arm / Rigid Hole}}\\
     & Large Cir. & 7.8 $\pm$ 1.65 & 62.3 $\pm$ 8.74 & - & 6/12\\ 
     & Pear & 11.0 $\pm$ 6.76 & 76.4 $\pm$ 28.10 & - & 5/12\\ 
     & Rectangle & 12.6 $\pm$ 4.37 & 92.3 $\pm$ 24.95 & - & 3/12\\

     & \multicolumn{5}{c}{\textbf{Rigid Hand / Rigid Arm / Compliant Hole}}\\
    & Large Cir. & 1.4 $\pm$ 0.12 & 65.6 $\pm$ 3.53 & - & 9/12 \\
     & Pear & 1.5 $\pm$ 0.13 & 68.9 $\pm$ 3.92 & - & 4/12 \\
      & Rectangle &1.5 $\pm$ 0.12 & 71.3 $\pm$ 4.35 &-& 12/12\\
      
    & \multicolumn{5}{c}{\textbf{Rigid Hand / Rigid Arm / Rigid Hole}}\\
     & Large Cir. & 2.2 $\pm$ 0.37 & 88.9 $\pm$ 10.48 & - & 6/12 \\
      & Pear & 2.2 $\pm$ 0.25 & 91.4 $\pm$ 8.46 & - & 0/12 \\
      & Rectangle & 2.1 $\pm$ 0.14 & 90.4 $\pm$ 13.09 & - & 2/12\\
     \hline
     
    \\ [-1.8ex]
    \multirow{8}{*}{\rotatebox[origin=c]{90}{\textbf{Limiting Control}}} 
     & \multicolumn{5}{c}{\textbf{Naive (omit Alg. \ref{alg:manipulate})}}\\
     & Large Cir. & 4.8 $\pm$ 1.78 & 78.0 $\pm$ 22.64 & - & 2/12\\ 
     & Pear & 7.7 $\pm$ 1.25 & 81.9 $\pm$ 10.13 & - & 0/12\\
     & Rectangle & 8.9 $\pm$ 0.95 & 103.5 $\pm$ 5.78 & - & 0/12\\ 
     
     & \multicolumn{5}{c}{\textbf{Open Loop (omit Alg. \ref{alg:manipulate} and Alg. \ref{alg:servo}})}\\
     & Large Cir. & 3.8 $\pm$ 1.09 & 53.47 $\pm$ 9.78 & - & 0/12\\ 
     & Pear & 5.1 $\pm$ 1.62 & 61.96 $\pm$ 4.01 & - & 0/12\\ 
     & Rectangle & 5.1 $\pm$ 1.22 & 65.45 $\pm$ 5.18 & - & 0/12\\ 
     \hline 
     
    \\ [-1.8ex]
    \multirow{8}{*}{\rotatebox[origin=c]{90}{\textbf{Noisy Sensing}}} 
     & \multicolumn{5}{c}{\textbf{Uniform Noise -- 5mm / 5$\degree$}}\\
     & Large Cir. & 5.9 $\pm$ 0.99 & 127.5 $\pm$ 18.11 & 50.6 $\pm$ 18.23 & 7/12\\ 
     & Pear & 10.7 $\pm$ 2.29 & 138.9 $\pm$ 12.46 & 40.2 $\pm$ 7.99 & 4/12\\
     & Rectangle & 9.9 $\pm$ 1.70 & 137.1 $\pm$ 6.28 & 49.7 $\pm$ 4.19 & 4/12\\
     
     & \multicolumn{5}{c}{\textbf{Uniform Noise -- 10mm / 10$\degree$}}\\
     & Large Cir. & 11.7 $\pm$ 2.29 & 148.5 $\pm$ 12.46 & 39.6 $\pm$ 11.19 & 1/12\\ 
     & Pear & 12.6 $\pm$ 0.33 & 134.1 $\pm$ 8.82 & 34.9 $\pm$ 7.19 & 0/12\\ 
     & Rectangle & 13.4 $\pm$ 1.97 & 212.6 $\pm$ 13.58 & 51.9 $\pm$ 11.79 & 0/12\\ 
     \hline 
    \end{tabular}
\end{table}
\vspace{-.1in}

\subsection{System Analysis} \label{sec:ablation}
Several ablation studies were performed using three of the five evaluated objects -- the large circle, the pear, and the rectangle -- in order to better understand how different components of the system contribute to task success. In particular, these tests evaluate the effects of: 1) reducing system compliance via the hand, the arm and the environment; 2) performing insertion with differing levels of feedback, i.e., naive and open loop control; and, 3) deliberately introducing noise into the object tracker's pose estimation to simulate higher perception uncertainty. (Table \ref{tab:ablations}). The \hyperref[sec:system recovery]{Appendix} describes how the system compensated for disturbances. 

 \begin{figure*}
 \centering
 \includegraphics[width = 0.97\textwidth]{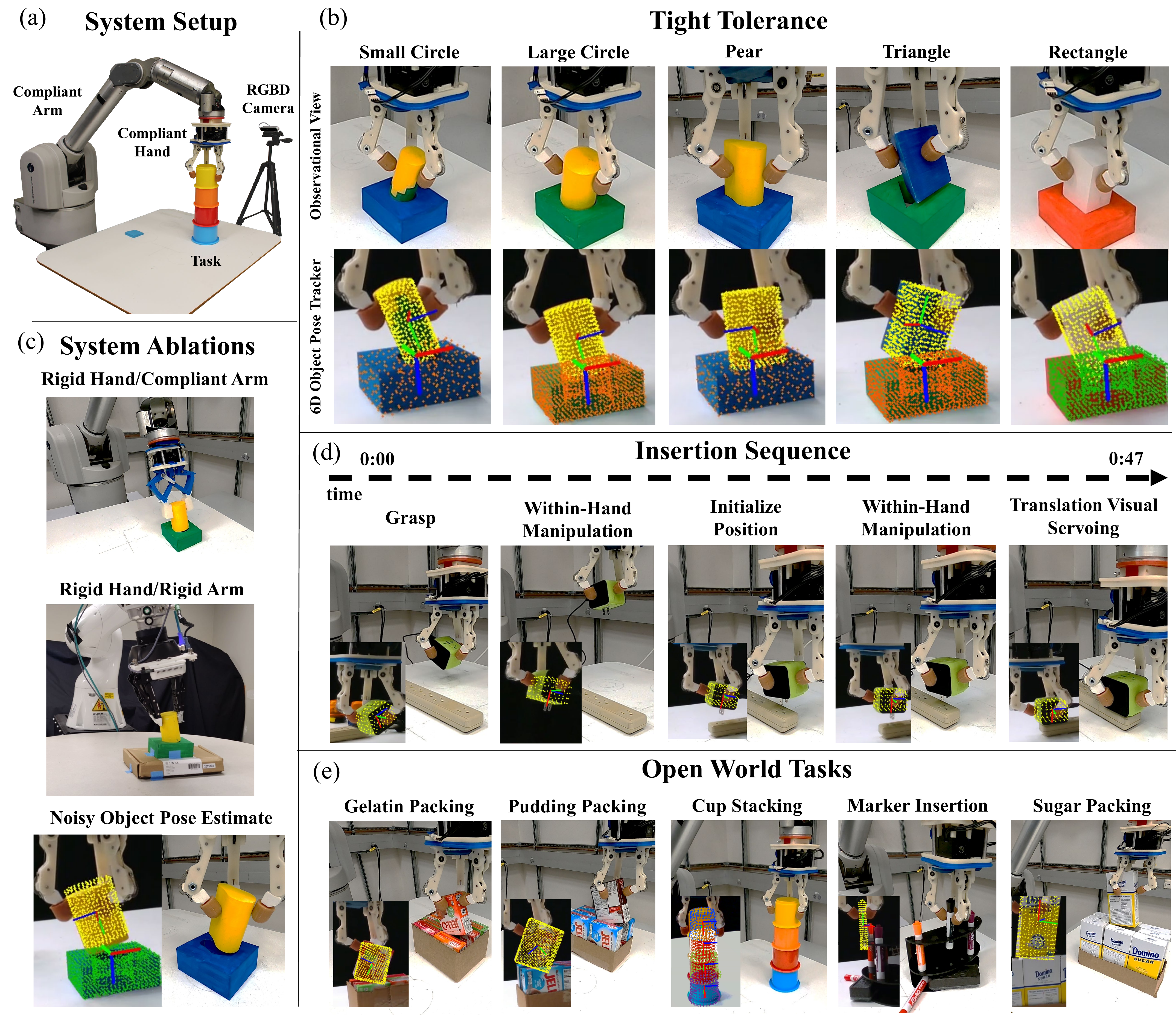}
 \caption{(a) System setup overview. (b) Tight tolerance insertion of 5 peg geometries, highlighting the observational/external view and the tracked 6D pose via the RGBD camera. (c) System ablations include reducing compliance of the hand and the arm, in addition to deliberately adding noise to the pose estimate. (d) The open world task of plug insertion is highlighted, showcasing the sequence of actions taken from grasp to insertion. (e) Five other open world tasks were also evaluated -- box packing, marker insertion, and cup stacking. Please refer to the supplementary video for a complete overview of evaluated tasks. }
 \vspace{-.2in}
 \label{fig:open_world_insertion} 
 \end{figure*}

\textbf{Reducing system compliance:} To test the role of compliance for task success, three altered system configurations were developed: 1) A system with a rigid, parallel Yale Openhand Model GR2 gripper \cite{bircher2017} with custom fingertips as to immobilize the object to the end of a low-impedance manipulator; 2) A system with a rigid 3-fingered Robotiq hand affixed to the end of a rigid Kuka IIWA manipulator but allowing compliance in the environment by attaching a standard packing box at the base of the hole; and, 3) The same rigid robot setup with the Robotiq hand and the Kuka manipulator but with a fixed hole, removing any presence of compliance. Fig. \ref{fig:both_systems} in the \hyperref[fig:both_systems]{Appendix} highlights the different variations considered in terms of compliance. Twelve insertions for each of the three test objects were performed in each case while following Alg. \ref{alg:mainAlgorithm}. During task execution, all object rotations that were performed by the hand originally, were now controlled solely by the manipulator. Results indicate, as presented in Table \ref{tab:ablations}, that compliance at some level in the arm/hand/object system is beneficial for task completion. The worst success rate arises for a fully rigid system, task, and environment. The Kuka setup with a compliant hole performed well, especially for the difficult rectangle insertion task, assisted by the precision of the object tracker and the higher accuracy of the rigid Kuka manipulator. Planning time between systems varied due to differences in computational resources, and thus it is not directly compared.

\textbf{Naive and open loop control:} These tests evaluate the effects of relinquishing feedback by limiting the amount of information transferred from the tracker to the insertion algorithm. This is tested both via naive control and an open loop configuration of the system. Naive control attempted to perform the same sequence of actions as in Alg. \ref{alg:mainAlgorithm}, but did not use any within-hand manipulation of the object after grasping, i.e., visual servoing was purely translational with the manipulator and Alg. \ref{alg:manipulate} was not used. The compliant hand in this case effectively acted as a remote center of compliance \cite{choi2016vision,loncaric1987normal}. In open loop testing, a single plan was computed and executed from the starting grasp configuration up until object insertion without utilizing in-hand manipulation or any visual feedback during the task, i.e., neither Alg. \ref{alg:manipulate} or Alg. \ref{alg:servo} were used. Notably, both of these ablations performed very poorly in testing, as the imprecision of the manipulator and lack of a controlled object rotation to aid in insertion, drastically impeded task success (Table \ref{tab:ablations}). 

\textbf{Noisy pose estimate:} The final ablation experiment evaluated how the accuracy of pose estimation played a role in task success. While following the same sequence of actions as in Alg. \ref{alg:mainAlgorithm}, uniformly sampled noise was introduced into the pose output of the tracker at two different levels. The first test introduced uniform sampled noise within $5mm$ and $5\degree$ to the pose estimate. The second test introduced uniform noise within the range of $10mm$ and $10\degree$. As indicated by the results in Table \ref{tab:ablations}, the system was largely able to compensate for this noise at the $5mm/5\degree$ level, successfully completing 7 for the large circle, 4 for the pear, and 4 for the rectangle out of 12 executions. The additional noise from the 10mm/10$\degree$ test drastically decreased success. The total task time for both noise level increased significantly as compared to the baseline trials in Sec. \ref{sec:tighttolerance}. This increase in execution time is attributed to Alg. \ref{alg:servo}, where noise continually moves $\mathcal{M}$ outside of the $\gamma$ threshold insertion region, so the manipulator slowly oscillates back and forth until $\delta_c$ is achieved. Conclusively, the ability of the algorithm to complete sub-$mm$ accuracy insertion tasks with purposefully added noise indicates the robustness of the overall framework.

\subsection{Open World Tasks} \label{sec:open world tasks}
Finally, a series of open world tasks were attempted to highlight the utility of the proposed system in complex manipulation scenarios: marker insertion, plug insertion, box packing, and cup stacking (Fig. \ref{fig:open_world_insertion}(e)). These tests mostly utilize objects contained in the YCB Object and Model Set \cite{calli2017}, which provides object CAD models for tracking. The charging plug, however, was 3D scanned using \cite{izadi2011kinectfusion} with an inexpensive RGBD sensor, providing a coarse representation of the true object model (Fig. \ref{fig:objects}). The goal for each task was to grasp, manipulate, and insert the object so as to reach its desired hole configuration, which was predefined depending on task requirements. Of the six tasks, three -- gelatin box, pudding box, and sugar box -- came in the form of packing, where a single box was removed from the case and had to be replaced. Two other tasks -- marker insertion and plug insertion -- were performed such that the marker was placed into a holder and the plug was inserted into an outlet. The most difficult of the open world tasks corresponded to cup stacking. This task requires sequential tracking of both the cup that is to be stacked on top of and the cup being manipulated. The proposed system is able to complete this task placing four cups successfully on top of one another. These evaluations showcase the tracking and manipulation capabilities of the proposed system, which are also highlighted in the supplementary video. 



\section{Discussions and Future Work} \label{sec:discussions and future work}

This manuscript presented a vision-driven servoing framework that tackles the problem of controlling compliant, passively adaptive mechanisms for precision manipulation. The framework is able to perform 5 tight tolerance and 6 open world tasks -- regardless of whether the object CAD model was imperfect or if pose estimate feedback was noisy (Sec. \ref{sec:open world tasks},\ref{sec:ablation}). In summary, the contributions of this work are fourfold:
\begin{itemize}
  \item \textbf{Precision control with minimal on-board sensing:} The framework can reliably complete an array of insertion tasks -- including those with tight tolerances -- without force-torque sensors on the manipulator or tactile sensors/joint encoders on the compliant hand. 
  \item \textbf{Passively compliant within-hand manipulation for insertion:} The system utilizes controlled hand actions to extend the task's workspace and provides an added layer of compliance to limit object insertion jamming.  
  \item \textbf{Vision-driven feedback controller:} The servoing insertion algorithm is able to generalize to different object geometries, and can be easily utilized for other robot assembly and insertion tasks given compliance.  
  \item \textbf{Utilizing the environment to solve the task:} The control strategy intentionally leverages premature contacts by relaxing the rigid constraints of the task, which is possible and effective given compliant systems. In this way, this work applies the principle of "extrinsic dexterity" \cite{Chavan-Dafle-2014-7860} in the context of insertion tasks with tight tolerances. 
\end{itemize}
 
There are several aspects of the proposed system that are worth pursuing in future work: 1) Adapt the insertion algorithm so as to better leverage compliance for complex out-of-plane geometry, such as in cases of sharp edges or non-convex objects; 2) Optimize a passively compliant hand design that is capable of performing both finger gaiting and in-hand manipulation with a large workspace; 3) Develop and integrate a model-free perception tracker into the visual feedback framework to reduce dependence on the object's CAD model; 4) Incorporate learning into the insertion algorithm so as to first automatically tune any object-specific hyperparameters and then incrementally develop more effective strategies in a RL fashion; and finally, 5) Increase difficulty of the tasks by performing more advanced and sequential assembly procedures, such assembling toys or simple pieces of household furniture.



\renewcommand*{\bibfont}{\footnotesize}
\bibliographystyle{IEEEtran}
\bibliography{INSERTION}

\clearpage
\begin{figure*}[t]
\centering
\includegraphics[width = \textwidth]{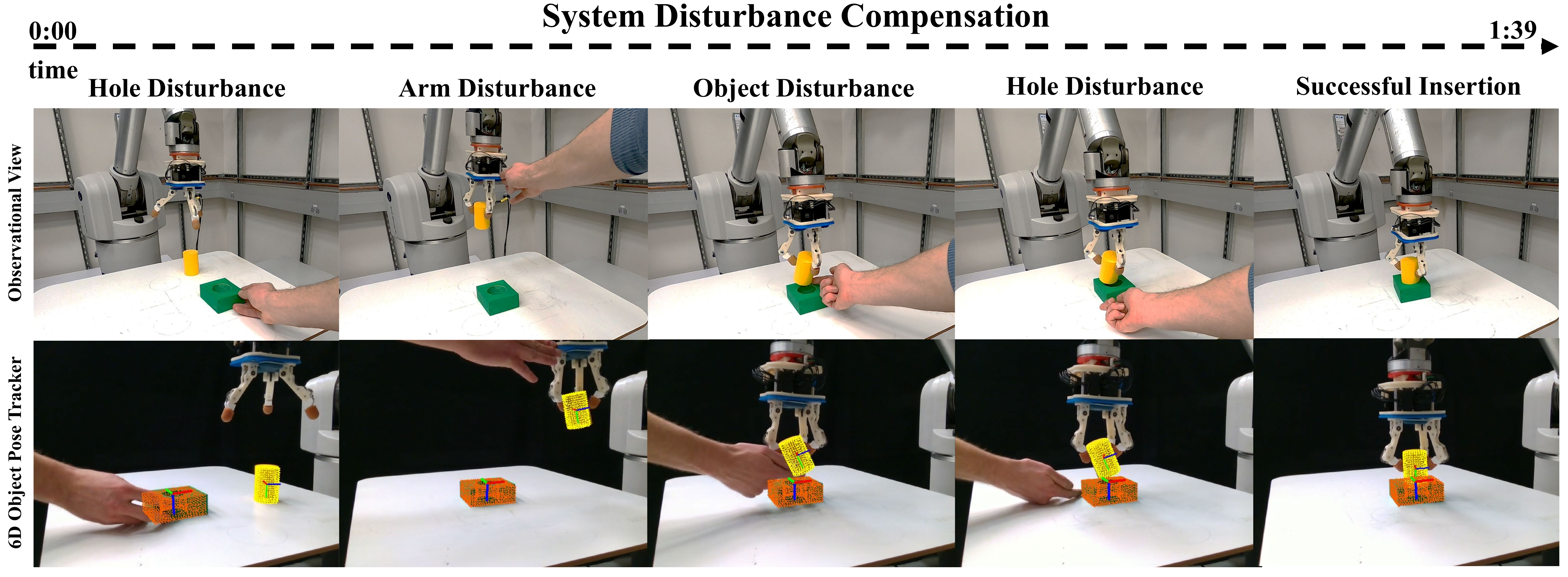}
\caption{The proposed system is robust to external disturbances imposed during task execution. In this task sequence, the pose of the hole, arm, and object are all manually perturbed and the system recovers such that the object successfully reaches its goal configuration. Refer to the supplementary video for the complete results.} \vspace{-.2in}
\label{fig:disturbance} 
\end{figure*}

\begin{figure}
\includegraphics[width = 0.48\textwidth]{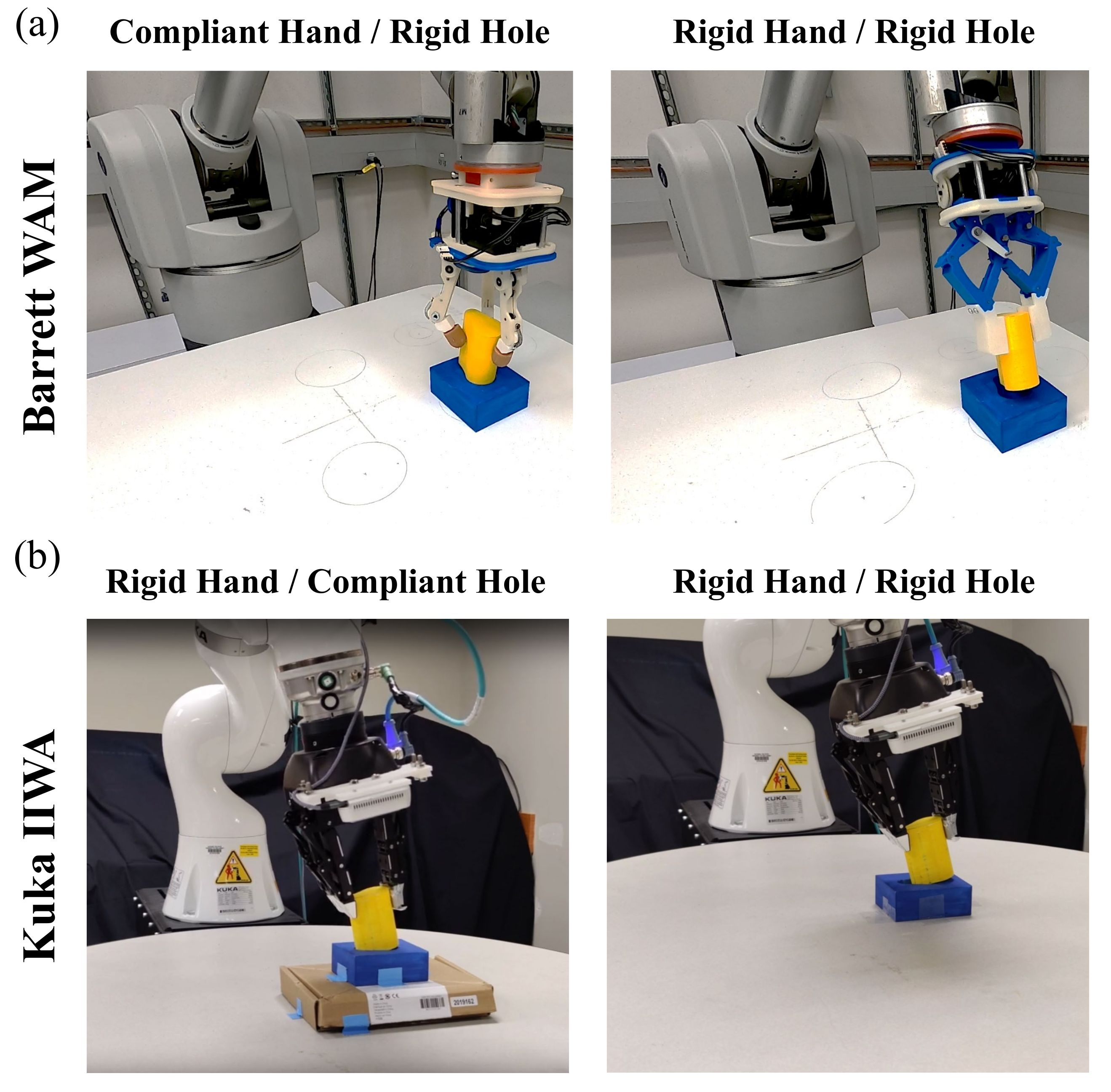}
\caption{The ablations make use of two manipulators with different levels of compliance at the arm, hand, or environment level. (a) A Barrett WAM serves as a low-impedance, compliant manipulator and has been tested with a compliant (left) and a rigid (right) hand; while the (b) Kuka IIWA is an example of a rigid manipulator, which has been tested with a rigid hand and a compliant (left) or rigid (right) hole.}

\label{fig:both_systems} 
\vspace{-.2in}
\end{figure}

\section*{Appendix} \label{sec: appendix}

\subsection{Selecting a Manipulation Frame} \label{sec:selecting a manipulation frame}
Selecting the best manipulation frame for object insertion, as eluded to in Sec. \ref{sec:condition for insertion}, is important as the antipodal grasp that defines $\mathcal{M}$, limits the amount of torque that can be applied by the contacts onto the object. Let's consider two objects that define different antipodal contact widths, $d_o^1$ and $d_o^2$, respectively. These distances fundamentally define the lever arm by which contact forces can be applied by the manipulator, or in the case of this work a hand, to the object. Assuming that friction coefficients and contact force applications are identical, the grasp with max($d_o^1, d_o^2$) will have a larger bound in the torques than can be imparted onto the object to aid in insertion.

In addition to the case of supplying a larger torque to the object, selecting the cross section of $\Gamma$ that nearly maximizes $d_o$ is also advantageous, according to the defined condition of insertion (Sec. \ref{sec:condition for insertion}). Fundamentally, while keeping the tolerance between the hole and all candidate object cross sections constant, as one increases $d_o$, $\beta_f$ will subsequently decrease. By selecting $d_o$ such that it is (near) maximized, this condition is  attempting to control the most difficult axis of the object to insert while relying on compliance to account for any out-of-plane reorientation of the other axis, which has a larger $\beta_f$. While the proposed approach for picking $\mathcal{M}$ is largely heuristic, it suits well for insertion with a dexterous hand that can apply much smaller forces than that of a manipulator.     

More formally, in the case of selecting $\mathcal{M} \in SE(3)$, we want to choose an insertion plane inside of an arbitrary 3D object geometry, $\Gamma$, such that we maximize the projection of $T$ back onto the vector connecting the two antipodal contact points. Assuming a given $\Gamma$, we can compute $T$ and thus $\mathcal{M}$ by analyzing properties of the object's face. More formally, consider that the outer geometry of a peg's face is sufficiently represented by a 2D point cloud of $n$ points, i.e. $F \in \rm I\!R ^ {n \times 2}$. It is possible to calculate the centroid of the point cloud, $F_c$, by taking its dimension-wise mean. Using Principal Component Analysis (PCA), we can compute the two principal axes of $F$, forming $P = [\pi_1, \pi_2]$. We solve for the outermost point, $m$, of the point cloud geometry by solving the optimization problem, \vspace{-0.1in}
\begin{equation} \label{PCA}
     m = \arg\max_{f\in F} \pi_1 \cdot{} (f-F_c) \vspace{-0.1in}
 \end{equation}
The location of $m$, which lies, or almost lies, along $\pi_1$, appropriately maximizes the distance, $d_o$, between two antipodal contacts on the 3D object geometry. We can then compute the transformation $T$ from $m$ back to the object frame, $\mathcal{X} \in SE(3)$, and thus providing our manipulation frame $\mathcal{M} \in SE(3)$ that will be used for guiding object insertion.  

\subsection{System Recovery to Disturbances} \label{sec:system recovery}
For robots that act in unknown, and contact-rich environment, various forms of disturbances can arise depending on the robot's operating scenario. As aforementioned, these disturbances can be self-induced, where for instance, a robot's perception and control noise causes the manipulated object to move differently than the internal model predicted, and thus the system must react. Similarly, disturbances can occur that are not caused directly by the robot's actions but by other objects or robots in its environment, as in the case of highly cluttered scenarios. We simulate such occurrences by deliberately disturbing the arm, object, and hole during an execution (Fig. \ref{fig:disturbance}). More specifically, during the task and in four separate occasions, we move the location of the hole on the support surface, causing our system to reactively adapt to the new goal configuration via the continuous feedback from the object tracker. During this sequence, we also push on the arm and the object, perturbing the state of the hand-object configuration, as to require the system to overcome such disturbances for successful insertion. These task evaluations are included in this paper's supplementary video.     

\subsection{System Observations and Limitations} \label{sec: limitations}
This subsection is part of a discussion on system limitations (Sec. \ref{sec:discussions and future work}), and what can be improved about the system for future advancements in robot assembly tasks. 

\textbf{Visual Perception}: While the RGBD-based 6D pose tracker is robust to a variety of objects with challenging properties such as textureless, reflective, geometrical featureless, or thin/small shapes, it struggles to track severely shiny, glossy, or transparent objects, due to the degenerated depth sensing of the camera. In future work, we hope to explore extending this framework to these other types of scenarios with the techniques of depth enhancement and completion \cite{sajjan2020clear}. In addition, the current framework requires an object CAD model beforehand to perform 6D pose tracking and for reasoning about the task of peg-in-hole insertion. In future work, reconstructing the model of novel objects on-the-fly \cite{mitash2020task} while with sufficient precision to perform high-tolerance tight insertion tasks is of interest.

\textbf{Inaccuracy of the Low-Impedance Manipulator for Grasp Acquisition: }While the manipulator leveraged in this work was largely beneficial for task completion, it also introduced difficulties in acquiring an initial grasp. Soft, compliant, and underactuated hands are well suited for grasping, but in cases of robot assembly where future within-hand manipulation is necessary (especially with 40+ finger actions as we saw with our tasks), acquiring a well-intended and stable grasp at the fingertips is necessary from the start. Such grasps would be especially possible with an accurate manipulator that is able to appropriately position the hand over the object for grasping. Though, for cases in this work where we saw task failure, it was largely due to starting with a poor initial fingertip grasp, which was directly attributed to the end effector's deviation from its commanded initial pose.   

\section*{Acknowledgements} \label{sec: acknowledgements}
The authors would like to thank V. Patel for his assistance in designing object-forming fingertips for the GR2 hand, in addition to the layered two-tone object geometries.

\end{document}